\pdfoutput=1
\documentclass{article}

\PassOptionsToPackage{numbers, sort&compress}{natbib}

\usepackage[main, final]{neurips_2026}

\usepackage[utf8]{inputenc}
\usepackage[T1]{fontenc}
\usepackage{hyperref}
\usepackage{url}
\usepackage{booktabs}
\usepackage{amsfonts}
\usepackage{nicefrac}
\usepackage{microtype}
\usepackage{xcolor}

\usepackage{multirow}
\usepackage{amsmath}
\usepackage{amssymb}
\usepackage{graphicx}
\usepackage{colortbl}
\usepackage{wrapfig}
\usepackage{algorithm}
\usepackage{algpseudocodex}
\usepackage{subcaption}
\usepackage{adjustbox}
\usepackage{longtable}
\usepackage{pdflscape}

\usepackage{amsthm}
\usepackage{makecell}

\newtheorem{proposition}{Proposition}

\newcommand{\method}{\texttt{DerivOpt}}
\newcommand{\methodbase}{\texttt{DerivBase}}
\newcommand{\bestsingle}{\texttt{BestSingle}}
\newcommand{\exprrel}{\mathrm{expr}_{L_2}}
\newcommand{\finerel}{\mathrm{hf}_{L_2}}
\newcommand{\qfine}{Q_{\mathrm{fine}}}
\newcommand{\eout}{e_{\mathrm{out}}}
\newcommand{\nrmse}{\mathrm{nRMSE}}
\newcommand{\horizon}{H}

\definecolor{myframepink}{RGB}{72,138,176}

\hypersetup{
    colorlinks=true,
    allcolors=myframepink
}

\title{Derived Fields Increase the Recoverable Information in Budgeted Neural PDE Surrogates}

\author{
Wenshuo Wang$^{1}$, Fan Zhang$^{2,3}$\thanks{Corresponding author.} \\
$^{1}$ School of Future Technology, South China University of Technology \\
$^{2}$ State Key Laboratory of Ocean Sensing \& Ocean College, Zhejiang University \\
$^{3}$ Kavli Institute for Astrophysics and Space Research, Massachusetts Institute of Technology \\
\texttt{202364870251@mail.scut.edu.cn}, \texttt{f.zhang@zju.edu.cn}
}

\begin{document}

\maketitle
\begingroup
\renewcommand{\thefootnote}{}
\footnotetext{Our code is available at \url{https://github.com/VinceWangVW/DerivOpt}.}
\endgroup

\begin{abstract}
For neural surrogates of time-dependent partial differential equations (PDEs), storage and memory budgets require training data to be obtained by coarsening and quantizing very-high-resolution numerical simulations. Fine-scale information can thus be lost before it reaches the model. We analyze this effect in periodic incompressible Navier--Stokes: quantization spreads error evenly over the retained modes, so the finest retained band, which holds the least energy, is the most corrupted, and velocity and vorticity are distorted differently under the same operator. Building on this analysis, we propose \textbf{Derived-Field Optimization (DerivOpt)}, which chooses which primitive and derived fields to carry from a supplied library, and how many bits each receives, by exactly minimizing a closed-form posterior risk under a calibrated channel model. This optimization problem always has an optimum and requires no knowledge of the governing equation. On \textsc{PDEBench}, DerivOpt not only lowers the pooled rollout \(\mathrm{nRMSE}\) but also substantially reduces the fine-band error. These gains are already present in the predictor-free input state and add to those of a multiscale architecture. Which fields are carried, and at what precision, is thus an optimizable design variable upstream of the predictor that determines how much information a fixed budget leaves recoverable.
\end{abstract}

\section{Introduction}\label{sec:intro}

Neural surrogates for time-dependent partial differential equations (PDEs) learn to advance the discretized state of a simulation by one step and are rolled out autoregressively in place of a numerical solver, at a fraction of its cost~\cite{brandstetter2022mpnpde,li2020fno,pfaff2020meshgraphnets,kochkov2021mlcfd,lippe2023pderefiner}. Existing neural simulators typically commit in advance to one state representation: primitive physical variables, hand-crafted derived variables, or latent coordinates~\cite{lino2023currentemerging,fukagata2025compressing}; see Section~\ref{sec:related}. Whatever the representation, training data and activations must fit storage and GPU memory, so high-precision, high-resolution simulation data are usually coarsened and quantized before training. On a given GPU, the budget for the fields on which a neural PDE surrogate can be trained is fixed. If the state representation is not treated as fixed, this raises a structural question that is largely orthogonal to the choice of backbone: under a fixed storage budget, which physical fields should be carried, and how should the budget be split among them?

To examine this question analytically and empirically, we use periodic incompressible Navier--Stokes as the clearest testbed and develop the canonical mechanism in Sections~\ref{sec:primitive}--\ref{sec:derivopt}. Section~\ref{sec:primitive} first shows that, under a fixed coarse operator and storage budget, modes above the retained cutoff are removed while quantization adds error evenly over the retained modes; because the finest retained band holds the least energy, its relative error is the largest, so the primitive state can already be detail-poor before it enters the model. Although they hold little energy, fine scales often govern the quantities that matter most in practice, such as peak local loads, front sharpness, mixing efficiency, boundary stress, and transport bottlenecks~\cite{park2016spacetime,yerragolam2022smallscale,snarski1995wallpressure}. Section~\ref{sec:fielddist} then shows that this fine-band distortion is not merely a generic quantization effect: under the same operator, primitive and derived fields suffer systematically different retained-band distortions, so some carried states preserve fine-scale information more faithfully than others. Building on this observation, Section~\ref{sec:derivopt} introduces \textbf{Derived-Field Optimization (\method{})}, a budgeted state-design framework that treats the carried physical fields as optimization variables. Once a family-specific analysis basis, a calibrated channel model, and a task-aligned target are given, it assigns every feasible field subset and bit allocation an explicit closed-form score and finds the optimum exactly. Notably, the problem is PDE-agnostic, and building the candidate library requires no knowledge of the governing equation.

The empirical scope is broader than the canonical Navier--Stokes derivation. The same state-design problem arises across benchmark PDE families with different boundary conditions, field semantics, and rollout regimes, so Section~\ref{sec:empirical} instantiates \method{} on the full time-dependent forward subset of \textsc{PDEBench}: advection, Burgers, diffusion--sorption, diffusion--reaction, radial dam break, incompressible Navier--Stokes, and compressible Navier--Stokes~\cite{takamoto2022pdebench}. Across the seven families, four backbones, and nine budget--resolution regimes (252 configurations, five seeds), \method{} gives the lowest pooled rollout \(\mathrm{nRMSE}\) in our comparison set, together with lower fine-band error and a longer detail-valid horizon. In incompressible Navier--Stokes, it lowers the predictor-free input fine-band error by 30\%, so more of the fine scales remain recoverable at the same budget. The rollout gains are largest at the tightest budget, vanish once the budget suffices for the primitive state, and add to those of a multiscale architecture.

Section~\ref{sec:discussion} returns to this broader perspective and argues that, in tightly budgeted neural simulation, the carried state determines how much information the budget leaves recoverable, whatever the predictor. Carried-state design should therefore be treated as a first-class design axis alongside architecture, loss, and rollout strategy, rather than as a fixed backdrop to predictor design.

\section{Related Work}\label{sec:related}

\subsection{Neural PDE surrogates across primitive, derived, and latent representations}

Existing neural simulators for time-dependent PDE forecasting can be organized not only by architecture, but also by the state representation they advance in time. Broadly, current work advances three kinds of representations: primitive physical fields, manually derived physical fields, and constructed reduced coordinates such as learned latent spaces or modal variables~\cite{lino2023currentemerging,fukagata2025compressing}.

One major paradigm advances primitive or directly observable variables. In this setting, neural simulators forecast future velocity-, pressure-, density-, or temperature-carrying states directly on grids or meshes~\cite{lee2019cylinder,pfaff2020meshgraphnets,qiu2024pifusion}. Another paradigm advances manually derived physical variables. Here the state is reparameterized into fields that expose incompressibility, transport, or rotational structure more explicitly, such as vorticity, potential vorticity, or Helmholtz potentials~\cite{li2020fno,kim2024decayingturbulence,lguensat2019qg,xing2023helmfluid,asaka2024vorticitywavelet}. A third paradigm performs rollout in compressed coordinates. In these models, the high-dimensional flow is first encoded into low-dimensional latent or reduced coordinates, and temporal dynamics are learned in that space before decoding back to the physical domain~\cite{wiewel2019latentspace,eivazi2020nonlinearmor,mohan2019compressedconvlstm,nakamura2021minimalchannel,solerarico2024betavae}.

Taken together, the dominant pattern in the literature is to commit to one fixed representation family before learning begins. Under storage constraints, however, this choice is not merely a modeling convenience; it determines how much fine-scale information survives coarsening and quantization. What is largely missing is work that treats the carried physical fields and their budget allocation as optimization variables under a fixed budget. \method{} addresses this gap directly. In tight-budget regimes, that design choice determines how much fine-scale information the predictor can recover in the first place.

\subsection{Long-horizon and fine-scale fidelity in neural PDE surrogates}\label{sec:long}

Beyond the choice of state representation, another central line of work asks how neural simulators can maintain accurate autoregressive rollouts over long horizons without washing out, destabilizing, or misrepresenting fine-scale structure~\cite{vinuesa2022enhancingcfd,lino2023currentemerging,wang2025fdbench}.

Within this line, five broad baseline families are especially representative of how current work tries to preserve fine detail under long autoregressive rollouts. The first is direct primitive-state autoregression, which advances only the observable state and serves as the default starting point for most learned simulators~\cite{lee2019cylinder,pfaff2020meshgraphnets,qiu2024pifusion,li2020fno}. The second is explicit single-derived-state rollout, which asks whether one transparent derived channel such as vorticity, a Helmholtz component, or another family-specific field already repairs the primitive bottleneck~\cite{kim2024decayingturbulence,lguensat2019qg,xing2023helmfluid,asaka2024vorticitywavelet}. The third keeps the carried state fixed but changes the training signal through rollout-aligned multiscale or spectral supervision~\cite{lippe2023pderefiner,worrall2024spectralshaping}. The fourth builds multiscale structure directly into the predictor architecture~\cite{rahman2022uno,gupta2021multiwavelet,fortunato2022multiscale,cao2022bistride}. The fifth learns a compressed carried state end to end under an explicit storage budget, typically by coupling an encoder--decoder surrogate to latent or entropy-model compression~\cite{wiewel2019latentspace,mohan2019compressedconvlstm,solerarico2024betavae,balle2018hyperprior}.

Taken together, dominant remedies remain model-side, loss-side, inference-side, or learned-bottleneck-side. They improve how a state is propagated, regularized, stabilized, or compressed, yet an overlooked issue is that, at the coarse resolutions typically used by budgeted simulators, the carried state may already lose much of the fine-scale information before rollout even begins. Once that state is detail-poor, downstream improvements can still reduce plain rollout error, but they generally cannot recover fine-scale information that is no longer present. In this sense, the central bottleneck under tight storage constraints often lies upstream of the predictor. \method{} therefore targets a more fundamental design axis: instead of assuming a fixed primitive rollout state and then improving propagation, it optimizes how a fixed state budget is distributed across primitive and derived fields so that fine-scale information is retained more faithfully from the outset.

\section{Unrecoverable High-Frequency Information of Primitive Fields Under Coarsening and Quantization}\label{sec:primitive}

We first isolate an error that arises before any learned predictor acts. Sections~\ref{sec:primitive} and~\ref{sec:fielddist} use periodic incompressible Navier--Stokes, the most analyzable standard testbed, with the state construction of Appendix~\ref{app:operator}. Let $u$ be the fine-grid velocity in $d$ dimensions, with Fourier coefficients $\hat u(k)$ at wavevectors $k$, $N_f$ and $N_c$ the fine and coarse resolutions per axis, $r=N_f/N_c$ the coarsening ratio, and $k_c=N_c/2$ the coarse cutoff, and define the retained band $\mathcal B_{\mathrm{exp}}$ and its top band $\mathcal B_{\mathrm{hf}}$ as
\begin{equation}
\mathcal B_{\mathrm{exp}}:=\{k:|k|<k_c\},
\qquad
\mathcal B_{\mathrm{hf}}:=\{k:\gamma k_c<|k|<k_c\},
\qquad
0<\gamma<1.
\label{eq:sec2_bands}
\end{equation}
For the decoded primitive input $\tilde u$, i.e., at rollout step 0, the input fine-band error is
\begin{equation}
\mathrm{hf}_{L_2}(0)^2
:=
\frac{\sum_{k\in\mathcal B_{\mathrm{hf}}}\|\widehat{\tilde u}(k)-\hat u(k)\|_2^2}
{\sum_{k\in\mathcal B_{\mathrm{hf}}}\|\hat u(k)\|_2^2}.
\label{eq:sec2_hf_metric}
\end{equation}

Standard primitive-state construction, which builds low-resolution training data for neural PDE surrogates from high-resolution simulation data, applies the spectral projector $P_{\mathrm{exp}}$ onto $\mathcal B_{\mathrm{exp}}$, samples the result on the coarse grid, quantizes each stored component $j$ with $b_j$ bits, $\mathbf b=(b_1,\ldots,b_d)$, and decodes by coarse trigonometric interpolation. Because the projector removes every mode outside $\mathcal B_{\mathrm{exp}}$, no unresolved mode folds back, and every retained coefficient satisfies $\widehat{\tilde u}(k)-\hat u(k)=\hat\eta(k)$, where $\hat\eta$ is the quantization noise lifted to the retained band. The numerator of Eq.~\eqref{eq:sec2_hf_metric} is thus the noise energy in $\mathcal B_{\mathrm{hf}}$, and the denominator is the signal energy there.

Under the standard high-rate closure (noise that is zero-mean, uncorrelated with the signal and across components, and white over the coarse grid), a uniform $b_j$-bit quantizer clipped at $a$ standard deviations adds noise of variance $\frac{a^2}{3}2^{-2b_j}\sigma_j^2$ to a component of variance $\sigma_j^2$ and spreads it evenly over the $N_c^d$ coarse modes, so
\begin{equation}
\sum_{k\in\mathcal B_{\mathrm{hf}}}\mathbb E\|\hat\eta(k)\|_2^2
=
N_c^d\,\vartheta_{\mathrm{hf}}\sum_{j=1}^{d}\frac{a^2}{3}\,2^{-2b_j}\sigma_j^2,
\qquad
\vartheta_{\mathrm{hf}}:=\frac{|\mathcal B_{\mathrm{hf}}|}{N_c^{d}},
\label{eq:sec3_noise}
\end{equation}
where $\vartheta_{\mathrm{hf}}$ is the fraction of coarse modes in $\mathcal B_{\mathrm{hf}}$ ($\vartheta_{\mathrm{hf}}\approx\frac{\pi}{4}(1-\gamma^2)$ in 2D). The signal energy in $\mathcal B_{\mathrm{hf}}$ is the fraction $\rho_{\mathrm{hf}}$ of the retained energy $N_c^d\sum_j\sigma_j^2$; for a shell spectrum $E(\kappa)$ with $\kappa=|k|$,
\begin{equation}
\sum_{k\in\mathcal B_{\mathrm{hf}}}\mathbb E\|\hat u(k)\|_2^2
=
\rho_{\mathrm{hf}}(r,\gamma)\,N_c^d\sum_{j=1}^{d}\sigma_j^2,
\qquad
\rho_{\mathrm{hf}}(r,\gamma)=\frac{\int_{\gamma k_c}^{k_c}E\,d\kappa}{\int_{0}^{k_c}E\,d\kappa}.
\label{eq:sec3_signal}
\end{equation}
For isotropic flow, $\sigma_j=\sigma$, and the ratio of Eqs.~\eqref{eq:sec3_noise} and~\eqref{eq:sec3_signal} gives
\begin{equation}
\mathbb E\big[\mathrm{hf}_{L_2}(0)^2\big]
=
\frac{\vartheta_{\mathrm{hf}}\,a^2}{3d\,\rho_{\mathrm{hf}}(r,\gamma)}\sum_{j=1}^{d}2^{-2b_j}
\ge
D_q(B,r,\gamma)
:=
\frac{\vartheta_{\mathrm{hf}}\,a^2}{3\,\rho_{\mathrm{hf}}(r,\gamma)}\,2^{-\frac{2B}{d}},
\label{eq:sec2_master_lb}
\end{equation}
where $B=\sum_j b_j$ is the primitive budget in bits per retained site. By Jensen's inequality, the equal split attains the minimum over splits with the same $B$, which defines the top-band quantization distortion $D_q$. Equation~\eqref{eq:sec2_master_lb} states the mechanism: quantization spreads error evenly over all coarse modes, but the top retained band holds only the fraction $\rho_{\mathrm{hf}}$ of the energy, so its relative error exceeds that of the whole field by the factor $(\vartheta_{\mathrm{hf}}/\rho_{\mathrm{hf}})^{1/2}$ and reaches order one at tight budgets. Appendix~\ref{app:theory_primitive} gives the full derivation.

\begin{figure}[t]
    \centering
    \begin{subfigure}[t]{0.525\linewidth}
        \centering
        \includegraphics[width=\linewidth]{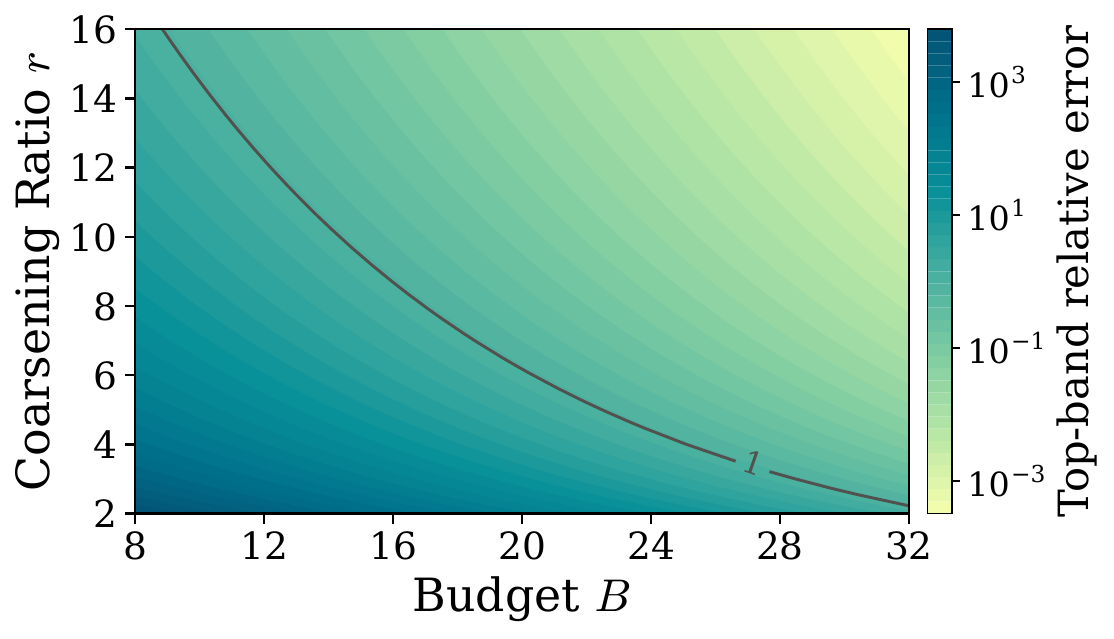}
        \caption{Top-band relative error $D_q^{1/2}$ of the decoded primitive velocity; log color scale, contour at 1. INS setting of Section~\ref{sec:empirical}, with the spectrum fitted to its measured direct-decode input error at $B=16$, $r=8$ (Appendix~\ref{app:theory_primitive}).}
        \label{fig:sec2_phase}
    \end{subfigure}
    \hfill
    \begin{subfigure}[t]{0.455\linewidth}
        \centering
        \includegraphics[width=\linewidth]{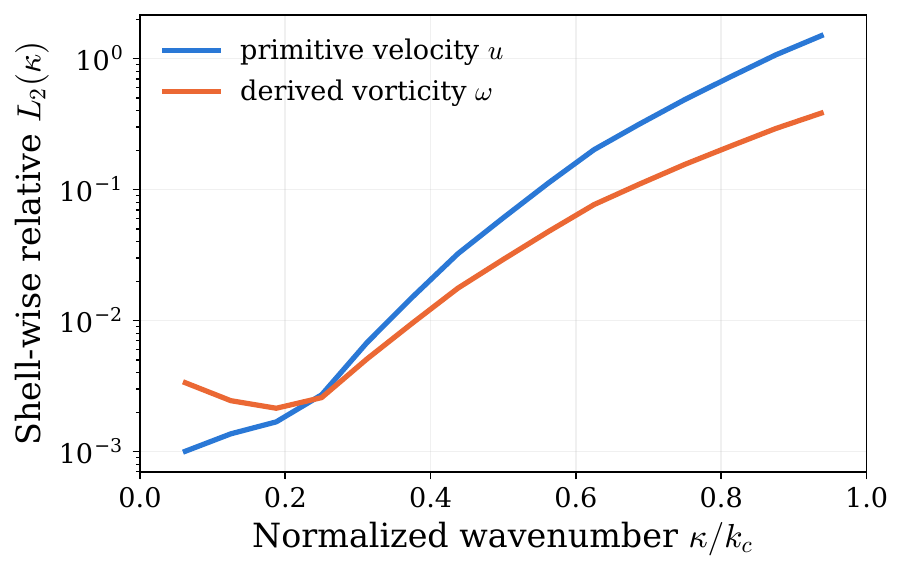}
        \caption{Shell-wise relative error of the latent coefficient reconstructed from $u$ alone or from $\omega$ alone under the same operator (Appendix~\ref{app:theory_fielddist}); $r=8$, 8 bits per stored component (equal bits per component, not equal storage), setting as in (a).}
        \label{fig:sec3_field_distortion}
    \end{subfigure}
\end{figure}

Figure~\ref{fig:sec2_phase} evaluates $D_q^{1/2}$ from Eq.~\eqref{eq:sec2_master_lb} over budgets $B$ and coarsening ratios $r$ in the INS setting of Section~\ref{sec:empirical}, computing $\rho_{\mathrm{hf}}$ at each $r$, with $k_c=N_f/(2r)$, from a shell spectrum fitted to the measured direct-decode input error at $B=16$, $r=8$. The error falls by a factor of two for each additional bit per velocity component, i.e., each increase of $B$ by 2, and exceeds one at small budgets, including the 2~KiB INS state ($B=16$, $r=8$). It depends on $r$ only through $\rho_{\mathrm{hf}}$: a larger $r$ lowers $k_c$ and moves the top band to more energetic wavenumbers, so $\rho_{\mathrm{hf}}$ grows and the error falls; a steeper spectrum, which lowers $\rho_{\mathrm{hf}}$, raises it at every budget. The error is fixed before prediction. Within the budgeted setting, a predictor, loss, or rollout scheme therefore starts from an input whose finest retained scales are corrupted, and it cannot reliably recover sample-specific detail that the state no longer contains.

\section{Field-Dependent Information Preservation}\label{sec:fielddist}

In this section, we find that the quantization sensitivity of the finest retained modes depends on which field is stored. In 2D incompressible flow, the divergence-free degree of freedom $z(k)$ satisfies $\hat u(k)=e_{\perp}(k)z(k)$ and $\hat \omega(k)=i|k|z(k)$, where $e_{\perp}(k)$ is the unit vector orthogonal to $k$. Each channel $c$ thus has a symbol $M_c$ with $\hat x_c(k)=M_c(k)z(k)$, namely $M_u(k)=1$ after projection onto $e_\perp(k)$ and $M_\omega(k)=i|k|$. Let $\hat z_c(k)$ be the latent coefficient reconstructed from channel $c$ alone. Under the operator of Section~\ref{sec:primitive}, the normalized per-mode distortion is
\begin{equation}
\mathcal D_c(k)
:=
\frac{\mathbb E|\hat z_c(k)-z(k)|^2}{S_z(k)}
=
\frac{\sigma^{q}_c(k;b_c)}{|M_c(k)|^2S_z(k)},
\label{eq:sec3_master_curve}
\end{equation}
where $S_z(k):=\mathbb E|z(k)|^2$ and $\sigma^{q}_c(k;b_c)$ is the quantization variance of channel $c$ at $b_c$ bits. For velocity and vorticity,
\begin{equation}
\mathcal D_u(k)
=
\frac{\sigma^{q}_u(k;b_u)}{S_z(k)},
\qquad
\mathcal D_{\omega}(k)
=
\frac{\sigma^{q}_{\omega}(k;b_{\omega})}{|k|^2S_z(k)}.
\label{eq:sec3_uomega_curves}
\end{equation}
Velocity pays the full quantization penalty on every shell; vorticity, mapped back to the latent coefficient, divides it by $|k|^2$ and is therefore better conditioned near the cutoff (Appendix~\ref{app:theory_fielddist}). To visualize this difference, Figure~\ref{fig:sec3_field_distortion} aggregates Eq.~\eqref{eq:sec3_master_curve} over shells at $r=8$ with 8 bits per stored component: velocity is less distorted on the low shells, vorticity near the cutoff, and the two curves cross at $\kappa\approx0.24\,k_c$. Different fields thus carry different parts of the spectrum well under the same operator. Which fields to carry, and how to split a fixed number of bits among them, is therefore a design problem.

\section{Derived-Field Optimization for Budgeted Neural Simulators}\label{sec:derivopt}

Sections~\ref{sec:primitive} and~\ref{sec:fielddist} show, in the canonical periodic incompressible Navier--Stokes setting, that under a fixed budget the simulator state itself should be designed rather than fixed \emph{a priori}. At the same time, the design rule is more general than this canonical specialization.

Let $\{\phi_{\ell}\}_{\ell\in\mathcal I}$ be an analysis basis adapted to the geometry and boundary conditions of a PDE family, $z(\ell)$ the latent coefficients of the state in that basis, and $\mathcal C$ a finite library of candidate fields. Each candidate $c$ has $m_c$ stored components per site and a linear relation $\hat x_c(\ell)=M_c(\ell)z(\ell)$, which is fitted on training data for candidates evaluated by physical operators (Appendix~\ref{app:calibration}). The library contains the primitive channels $\mathcal P$ and fields obtained from them by fixed operators, such as spectral powers of a boundary-adapted Laplacian, gradients, curl and divergence, or band projections (Appendix~\ref{app:libraries}). Building it uses the field types, the geometry, and the boundary conditions, but not the governing equation, its coefficients, or the identity of the PDE. Under a shared budgeted coarsen--quantize--decode operator (Appendix~\ref{app:coarsen_protocol}), each decoded channel is modeled as
\begin{equation}
\hat y_c(\ell)=B_c(\ell)M_c(\ell)z(\ell)+n_c(\ell),
\label{eq:sec4_channel}
\end{equation}
where the in-band transfer $B_c(\ell)$ and the covariance $\Sigma_c(\ell;b_c)$ of the residual $n_c(\ell)$ are estimated from training data for every admissible bit width $b_c$, with full covariance across the stored components.

For a target $\hat g(\ell)=L_g(\ell)z(\ell)$ on an index set $\mathcal I_{\star}$, a design is a nonempty subset $S\subseteq\mathcal C$ of fields to carry together with their bit widths $\mathbf b=(b_c)_{c\in S}$, and $\hat g_S(\ell)$ denotes the target decoded from the stored channels of $S$. \textbf{Derived-Field Optimization} (\method{}) chooses the design that recovers the target best on average within a budget of $B$ bits per retained site:
\begin{equation}
\begin{aligned}
(S^{\star},\mathbf b^{\star})
&\in
\arg\min_{(S,\mathbf b)\in\mathcal D_B}
\sum_{\ell\in\mathcal I_{\star}}w_{\ell}\,\mathbb E\big\|\hat g_S(\ell)-\hat g(\ell)\big\|_2^2,\\
\mathcal D_B&:=\Big\{(S,\mathbf b):\ \emptyset\neq S\subseteq\mathcal C,\ \ b_c\in\mathbb Z_{\ge 1}\ \text{for}\ c\in S,\ \ \textstyle\sum_{c\in S}m_cb_c\le B\Big\},
\end{aligned}
\label{eq:sec4_problem}
\end{equation}
where $w_\ell\ge0$ are task weights and unused bits are zero padding. Stacking the selected channels gives the linear Gaussian model $\hat y_S(\ell)=A_S(\ell)z(\ell)+n_S(\ell)$, with $A_S(\ell):=\operatorname{rowstack}_{c\in S}[B_c(\ell)M_c(\ell)]$ and stacked residual covariance $\Sigma_S(\ell;\mathbf b)$. Under the prior $z(\ell)\sim\mathcal{CN}(0,S_z(\ell))$, the error-minimizing decoder is the posterior mean $\hat z_S(\ell)$, so $\hat g_S=L_g\hat z_S$; this posterior mean is also the shared decoder of every explicit-state method in Section~\ref{sec:empirical}. Its error is set by the posterior covariance $\Sigma_{\mathrm{post},S}$, so the objective of Eq.~\eqref{eq:sec4_problem} equals the design risk
\begin{equation}
\begin{aligned}
\mathcal J(S,\mathbf b)&=\sum_{\ell\in\mathcal I_{\star}}w_{\ell}\,\mathrm{tr}\Big(L_g(\ell)\,\Sigma_{\mathrm{post},S}(\ell)\,L_g(\ell)^{\ast}\Big),\\
\Sigma_{\mathrm{post},S}(\ell)
&=
\Big(S_z(\ell)^{-1}+A_S(\ell)^{\ast}\Sigma_S(\ell;\mathbf b)^{-1}A_S(\ell)\Big)^{-1}
\end{aligned}
\label{eq:sec4_posterior}
\end{equation}
(Appendix~\ref{app:theory_derivopt}). $\mathcal J$ is the expected weighted error of recovering the target from the stored bits, so its minimum over $\mathcal D_B$ measures how much of the target a budget $B$ leaves recoverable under the calibrated model. It involves only calibrated statistics and no predictor, so once the channels are calibrated every feasible design has a closed-form score, and Eq.~\eqref{eq:sec4_problem} reduces to minimizing this score over $\mathcal D_B$.

\begin{proposition}[Existence, exactness, and primitive fallback]\label{prop:existence}
Let $\mathcal C$ be finite and contain the nonempty set $\mathcal P$ of primitive channels, let $B\in\mathbb N$ with $B\ge\min_{c\in\mathcal P}m_c$, and let $S_z(\ell)\succ 0$. Then (i) $\mathcal D_B$ is finite and nonempty and $\mathcal J$ is finite on it, so a global minimizer $(S^\star,\mathbf b^\star)$ exists and finite search finds it exactly; (ii) $\mathcal J(S^\star,\mathbf b^\star)\le\min\{\mathcal J(S,\mathbf b):(S,\mathbf b)\in\mathcal D_B,\ S\subseteq\mathcal P\}$, with equality if and only if no design that carries a non-primitive channel has lower risk than the best primitive-only design; (iii) for libraries $\mathcal C\subseteq\mathcal C'$, the optimal risk over $\mathcal C'$ is at most that over $\mathcal C$.
\end{proposition}
\begin{proof}[Proof sketch]
Each $b_c$ lies in $\{1,\ldots,\lfloor B/m_c\rfloor\}$ and $\mathcal C$ is finite, so $\mathcal D_B$ is finite, and it is nonempty because the primitive channel with the fewest components fits the budget at one bit per component. Since $S_z^{-1}\succ0$ and $A_S^\ast\Sigma_S^{-1}A_S\succeq0$, the matrix inverted in Eq.~\eqref{eq:sec4_posterior} is positive definite, so $\mathcal J$ is finite, and a finite nonempty set attains its minimum. The primitive-only designs form a subset of $\mathcal D_B$, and $\mathcal D_B(\mathcal C)\subseteq\mathcal D_B(\mathcal C')$; a minimum over a superset cannot be larger. Appendix~\ref{app:proof} gives the details and an extended Navier--Stokes library.
\end{proof}

Proposition~\ref{prop:existence} makes \method{} applicable to any PDE family and any finite library for which Eq.~\eqref{eq:sec4_channel} can be calibrated. The problem is PDE-agnostic: its formulation, calibration, and solution use only the candidate library and training snapshots, never the governing equation. No winning field has to be specified in advance, and a library without useful candidates returns the best primitive-only design. The proposition concerns the calibrated risk; whether a lower risk yields better rollouts is tested in Section~\ref{sec:empirical}. We solve Eq.~\eqref{eq:sec4_problem} with an exact branch-and-bound search that prunes only by valid lower bounds on $\mathcal J$ and breaks ties toward lower cost, never by outcome metrics (Appendix~\ref{app:selection}).

\paragraph{Periodic incompressible Navier--Stokes.} Returning to the canonical example of Sections~\ref{sec:primitive} and~\ref{sec:fielddist}, the design risk can be written in closed form. With velocity as the target, incompressibility leaves $d-1$ latent degrees of freedom per mode and velocity is an isometry of them, so $\mathcal J=\sum_{k\in\mathcal B_\star}w(k)\operatorname{tr}\Sigma_{\mathrm{post},S}(k)$, where $\mathcal B_\star$ is the set of target modes. For $\mathcal C=\{u,\omega\}$ with isotropic residuals $\Sigma_c=\sigma_c^{\mathrm{eff}}I$, the channel precisions add, and the score becomes
\begin{equation}
\mathcal J_{\mathrm{NS}}(S,\mathbf b)
=
(d-1)
\sum_{k\in\mathcal B_{\star}}w(k)
\Bigg(
S_z(k)^{-1}
+
\mathbf 1_{\{u\in S\}}\frac{|B_u(k)|^2}{\sigma^{\mathrm{eff}}_u(k;b_u)}
+
\mathbf 1_{\{\omega\in S\}}\frac{|B_{\omega}(k)|^2|k|^2}{\sigma^{\mathrm{eff}}_{\omega}(k;b_{\omega})}
\Bigg)^{-1}.
\label{eq:sec4_ns_objective}
\end{equation}
The bracket is the posterior precision of mode $k$: the prior precision $S_z(k)^{-1}$ plus the calibrated signal-to-noise ratio contributed by each carried field, and $\mathcal J_{\mathrm{NS}}$ sums the resulting posterior variances over the target modes with weights $w(k)$. The vorticity term carries the extra factor $|k|^2$ of Section~\ref{sec:fielddist}, so its contribution grows toward the cutoff, whereas that of velocity does not. Each contribution costs $m_cb_c$ bits, and the optimizer trades the precision a field adds against these bits. Appendix~\ref{app:theory_derivopt} derives Eq.~\eqref{eq:sec4_ns_objective}, and Appendix~\ref{app:family_inst} gives the instantiations for the other families.

\section{Empirical Evaluation}\label{sec:empirical}

We test two claims in the budgeted setting. The mechanism claim is that the carried state is already detail-poor at input and that derived fields reduce this input error. The outcome claim is that this loss carries into rollouts, where predictor- or codec-side alternatives at the same budget do not recover it, whereas \method{} reduces it.

\subsection{Experimental setup}\label{sec:exp_setup}

\paragraph{Benchmark.} We use the seven time-dependent forward families of \textsc{PDEBench}~\cite{takamoto2022pdebench}: 1D advection, Burgers, and diffusion--sorption, and 2D diffusion--reaction, radial dam break, incompressible Navier--Stokes (INS), and compressible Navier--Stokes. They cover periodic and non-periodic boundaries and scalar and multi-field states. Each family uses 1,000 trajectories split 640/104/256 into training, validation, and test, with one test cohort shared by all methods, budgets, and seeds. \textsc{PDEBench} INS is wall-bounded; it is the closest benchmark counterpart of Sections~\ref{sec:primitive}--\ref{sec:derivopt} rather than an instance of the periodic analysis. Grids and time windows are listed in Appendix~\ref{app:protocol}.

\paragraph{Budget.} A configuration fixes a family, a backbone, a budget ratio $R\in\{0.25,0.5,1\}$, and a per-axis retained fraction $f\in\{1/8,1/4,1/2\}$, giving $7\times4\times3\times3=252$ configurations. The state passed between autoregressive steps keeps $n_r=(fN)^d$ sites, where $N$ is the reference resolution per axis, and may use $B=32\,m_pR$ bits per site, where $m_p$ is the number of primitive scalar components; that is, the fraction $R$ of FP32 primitive storage at that resolution. Every serialized bit counts against the cap $n_rB$, including field codes, latent codes, headers, and padding, whereas statistics fixed in the checkpoint cost nothing (Appendix~\ref{app:protocol}). The canonical regime $(R,f)=(0.25,1/8)$ was fixed on a training-split INS pilot before any test evaluation (Appendix~\ref{app:canonical}); for INS it gives $32^2$ sites at 16 bits per site, i.e., 2,048~B per state.

\paragraph{Methods.} All methods share one predictor interface: the network reads and predicts decoded coarse primitive fields, derived fields are never added as network inputs, and the external state passed between steps is the encoded payload. The state-design ladder consists of \texttt{Primitive} (primitive fields only, bits split evenly), \bestsingle{} (the best single derived field of the library, with all bits), \methodbase{} (the field subset selected by \method{}, with bits split evenly across its components and the remainder padded), and \method{}. All four use the shared posterior decoder, so differences within the ladder come from the carried state alone.

We also compare three strong baselines for high-frequency recovery at the same cap, one from each of the other three families of Section~\ref{sec:long}: \texttt{ArchMulti}, a multiscale variant of each backbone in the spirit of~\cite{rahman2022uno,gupta2021multiwavelet}; \texttt{RolloutMulti}, rollout-aligned multiscale supervision in the spirit of~\cite{lippe2023pderefiner,worrall2024spectralshaping}; and \texttt{LatentAE-Hyper}, a latent autoencoder~\cite{wiewel2019latentspace} with a hyperprior entropy model~\cite{balle2018hyperprior} whose realized bitstream is rate-controlled to the cap. \texttt{ArchMulti} and \texttt{RolloutMulti} carry the \texttt{Primitive} state. \method{} is run once per family and regime on training data (63 selector runs), with the libraries of Appendix~\ref{app:libraries}, the primitive field as the target, and weights $w_\ell$ fixed in advance at 0.35 on the low and middle bands and 0.65 on the top band, normalized within each group by training statistics.

\paragraph{Training.} The backbones are FNO, U-Net, ConvLSTM, and Transformer, each trained as a one-step autoregressive predictor. For every method and configuration, 12 hyperparameter trials (learning rate, weight decay, and gradient clipping), each of 10,000 updates, are compared on the validation split with seed 11; the selected setting is then trained with four further seeds (29, 47, 71, 103). We report means over the five seeds.

\paragraph{Metrics.} In each family's boundary-adapted basis, let $P_{\exp}$ project onto the retained band and $P_{\mathrm{fine}}$ onto its top band, $(0.75\,k_c,k_c)$. For the primitive target $x_h$ at step $h$ and its decoded or predicted estimate $\hat x_h$,
\begin{equation}
\finerel(h)=\frac{\|P_{\mathrm{fine}}(\hat x_h-x_h)\|_2}{\|P_{\mathrm{fine}}x_h\|_2},
\qquad
\horizon=\frac{1}{T}\max\big\{\tau\le T:\ \text{steps }1,\ldots,\tau\text{ are all valid}\big\}.
\label{eq:sec6_metrics}
\end{equation}
The input error $\finerel(0)$ is measured on the decoded input with no predictor. Over closed-loop rollouts that run to the end of each stored trajectory, $h=1,\ldots,T$ (Appendix~\ref{app:protocol}), $\nrmse$ and mean $\finerel$ sum squared errors and squared reference norms over the 256 test trajectories and all $T$ steps and take the square root of their ratio, for the full field and for the top band, respectively. The detail-valid horizon $\horizon\in[0,1]$ is computed per trajectory ($\horizon=0$ if step 1 is invalid) and averaged over trajectories. A step is valid when the prediction is accurate and spectrally faithful at once: $\exprrel(h)<1$ on the retained band, $\finerel(h)<1$, the fine-band energy ratio $\qfine(h)\in[0.8,1.25]$, and the out-of-band leakage $\eout(h)<0.1$ (definitions in Appendix~\ref{app:metrics}); one threshold set is used for all families, methods, and backbones. Pooled values average over configurations and then over seeds, and paired 95\% intervals use a 12,000-sample hierarchical bootstrap over family, configuration, and paired seed.

\subsection{Main results}\label{sec:results}

\subsubsection{Input-stage mechanism in the canonical case and across the full suite}\label{sec:mechanism}

\begin{table}[t]
    \centering
    \caption{INS in the canonical regime (lower is better). Carried states: \texttt{Primitive}, \texttt{ArchMulti}, and \texttt{RolloutMulti} $u$\,8+8; \bestsingle{} $\omega$\,16; \methodbase{} $u$\,5+5, $\omega$\,5 (one padding bit); \method{} $u$\,4+4, $\omega$\,8; \texttt{LatentAE-Hyper} a learned code padded to the cap. The encoder of \texttt{LatentAE-Hyper} is trained with each backbone, so its $\finerel(0)$ averages the four backbones within each seed.}
    \label{tab:ins}
    \setlength{\tabcolsep}{4pt}
    \resizebox{\textwidth}{!}{%
    \begin{tabular}{lccccccccc}
    \toprule
    & Input & \multicolumn{2}{c}{FNO} & \multicolumn{2}{c}{U-Net} & \multicolumn{2}{c}{ConvLSTM} & \multicolumn{2}{c}{Transformer} \\
    \cmidrule(lr){2-2}\cmidrule(lr){3-4}\cmidrule(lr){5-6}\cmidrule(lr){7-8}\cmidrule(lr){9-10}
    & & & mean & & mean & & mean & & mean \\
    Method & $\finerel(0)$ & $\nrmse$ & $\finerel$ & $\nrmse$ & $\finerel$ & $\nrmse$ & $\finerel$ & $\nrmse$ & $\finerel$ \\
    \midrule
    \texttt{Primitive}      & 0.999 & 0.0680 & 0.831 & 0.0749 & 0.855 & 0.1054 & 0.889 & 0.0943 & 0.893 \\
    \bestsingle{}           & 0.863 & 0.0661 & 0.755 & 0.0718 & 0.778 & 0.0989 & 0.800 & 0.0896 & 0.820 \\
    \methodbase{}           & 0.830 & 0.0644 & 0.682 & 0.0697 & 0.704 & 0.0971 & 0.727 & 0.0867 & 0.746 \\
    \method{}               & \textbf{0.700} & 0.0585 & \textbf{0.584} & \textbf{0.0634} & \textbf{0.597} & \textbf{0.0880} & \textbf{0.634} & \textbf{0.0785} & \textbf{0.630} \\
    \midrule
    \texttt{ArchMulti}      & 0.999 & \textbf{0.0582} & 0.685 & \textbf{0.0634} & 0.700 & 0.0883 & 0.720 & 0.0791 & 0.734 \\
    \texttt{RolloutMulti}   & 0.999 & 0.0675 & 0.794 & 0.0720 & 0.815 & 0.1008 & 0.820 & 0.0906 & 0.855 \\
    \texttt{LatentAE-Hyper} & 0.812 & 0.0633 & 0.672 & 0.0692 & 0.689 & 0.0962 & 0.716 & 0.0868 & 0.731 \\
    \bottomrule
    \end{tabular}%
    }
\end{table}

The first column of Table~\ref{tab:ins} removes the predictor. At the 2~KiB INS budget, the primitive state decoded by the shared posterior decoder has $\finerel(0)=0.999$: the error in the top retained band is as large as the signal, so the predictor receives essentially none of the information in this band. Decoding the same bits directly, without fusion, gives 1.394, so the shared decoder already helps and is not the source of the gap. With the decoder, the target, and the inputs fixed, changing only the carried state lowers $\finerel(0)$ to 0.863 with vorticity alone (\bestsingle{}), to 0.830 with velocity and vorticity at equal bits (\methodbase{}), and to 0.700 with \method{}'s allocation, a 29.9\% reduction. Field mixing and allocation both contribute, and only with the optimized allocation is the carried state more accurate than the learned code of \texttt{LatentAE-Hyper} at the same cap (0.812).

In this wall-bounded flow, vorticity alone is the most accurate in the interior but the least accurate near the walls (0.689 and 1.300, against 1.054 and 0.777 for \texttt{Primitive}), whereas \method{} is the most accurate near the walls (0.605) and overall, though not everywhere (Appendix~\ref{app:extra}); each field is thus the more accurate one where the other is weak, and the mixed state combines them.

Across the suite (Table~\ref{tab:suite}b), \method{} lowers the pooled $\finerel(0)$ from 1.006 to 0.710 at $R=0.25$ and from 0.291 to 0.101 at $R=0.5$; at $R=1$ it selects the primitive encoding, and the input is unchanged. The input loss of the primitive state thus grows as the budget tightens, as Section~\ref{sec:primitive} predicts, and derived fields reduce it at both tighter budgets. Under a fixed budget, the fine-scale information available to the predictor is therefore set by the carried state before any learning, and carrying derived fields with an optimized bit allocation makes more of it recoverable from the same bits, except when the budget already suffices for the primitive state.

\begin{table}[t]
    \centering
    \caption{All 252 configurations (seed standard deviations are at most $1\times10^{-4}$ for $\nrmse$, $0.0022$ for mean $\finerel$, and $0.002$ for $\horizon$). (a)~Pooled over families, backbones, and regimes. (b)~Pooled by budget ratio $R$; \texttt{ArchMulti} carries the \texttt{Primitive} input, and at $R=1$ \method{} selects the primitive encoding. Lower is better except for $\horizon$.}
    \label{tab:suite}
    \small
    \begin{minipage}[t]{0.47\linewidth}
    \centering
    \textbf{(a)} Pooled\\[2pt]
    \setlength{\tabcolsep}{3.5pt}
    \begin{tabular}{lccc}
    \toprule
    Method & $\nrmse$ & \makecell{mean\\$\finerel$} & $\horizon$ \\
    \midrule
    \texttt{Primitive}      & 0.0673 & 0.764 & 0.384 \\
    \bestsingle{}           & 0.0678 & 0.743 & 0.375 \\
    \methodbase{}           & 0.0650 & 0.701 & 0.451 \\
    \method{}               & \textbf{0.0614} & \textbf{0.653} & \textbf{0.524} \\
    \midrule
    \texttt{ArchMulti}      & 0.0646 & 0.699 & 0.458 \\
    \texttt{RolloutMulti}   & 0.0664 & 0.742 & 0.410 \\
    \texttt{LatentAE-Hyper} & 0.0649 & 0.696 & 0.463 \\
    \bottomrule
    \end{tabular}
    \end{minipage}\hfill
    \begin{minipage}[t]{0.51\linewidth}
    \centering
    \textbf{(b)} By budget ratio\\[2pt]
    \setlength{\tabcolsep}{3.5pt}
    \begin{tabular}{llccc}
    \toprule
    & Method & $R=0.25$ & $0.5$ & $1$ \\
    \midrule
    \multirow{2}{*}{$\finerel(0)$} & \texttt{Primitive} & 1.0057 & 0.2913 & 0.0205 \\
                                   & \method{}          & \textbf{0.7099} & \textbf{0.1008} & 0.0205 \\
    \midrule
    \multirow{3}{*}{$\nrmse$}      & \texttt{Primitive} & 0.0735 & 0.0672 & 0.0611 \\
                                   & \texttt{ArchMulti} & 0.0672 & 0.0656 & \textbf{0.0610} \\
                                   & \method{}          & \textbf{0.0621} & \textbf{0.0609} & 0.0611 \\
    \midrule
    \multirow{3}{*}{$\horizon$}    & \texttt{Primitive} & 0.127 & 0.385 & 0.641 \\
                                   & \texttt{ArchMulti} & 0.316 & 0.404 & \textbf{0.653} \\
                                   & \method{}          & \textbf{0.446} & \textbf{0.484} & 0.641 \\
    \bottomrule
    \end{tabular}
    \end{minipage}
\end{table}

\subsubsection{Rollout accuracy in the canonical case and across the full suite}\label{sec:outcome}

The input deficit carries into rollouts. In INS (Table~\ref{tab:ins}), \method{} has the lowest mean $\finerel$ on all four backbones, 0.584--0.634 against 0.672--0.731 for the next best, \texttt{LatentAE-Hyper}, and on every backbone the four ladder states have the same order in mean $\finerel$ as in $\finerel(0)$: for a given backbone, a more accurate input gives a more accurate rollout. \method{} is also below \texttt{ArchMulti} and \texttt{RolloutMulti}, which change the predictor or the loss but receive the primitive input; a stronger predictor or a rollout-aligned loss improves on \texttt{Primitive} but does not make up for what the primitive input lacks. Its $\nrmse$ is close to that of \texttt{ArchMulti}, lower on three of the four backbones in the five-seed mean and in 12 of the 20 backbone--seed pairs, and its horizon $\horizon$ exceeds that of \texttt{ArchMulti} in all 20 pairs: the two are comparable on the full field, but \method{} stays detail-valid for longer.

Across the 252 configurations (Table~\ref{tab:suite}a), \method{} has the lowest pooled $\nrmse$, the lowest pooled mean $\finerel$ (0.653, against 0.696 for \texttt{LatentAE-Hyper}), and the highest $\horizon$. Its $\nrmse$ is 5.0\% below that of \texttt{ArchMulti} and 5.4\% below that of \texttt{LatentAE-Hyper}, the two strongest alternatives, with paired differences of $-0.0032$ $[-0.0053,-0.0014]$ and $-0.0035$ $[-0.0055,-0.0016]$ (lower in 162 and 161 of the 252 configurations), and its $\horizon$ is higher by $0.066$ $[0.037,0.096]$ and $0.061$ $[0.032,0.092]$. \method{} also improves pooled $\nrmse$ and $\horizon$ over \texttt{ArchMulti} in each of the seven families. The INS pattern thus holds across the suite, with a moderate gain in full-field error and a larger one in the detail-valid horizon. The ladder shows where the gain comes from. A single derived field lowers the mean $\finerel$ (0.743 for \bestsingle{} against 0.764 for \texttt{Primitive}) but not the pooled $\nrmse$ or $\horizon$, on which \bestsingle{} is slightly worse than \texttt{Primitive}. Mixing primitive and derived fields at equal bits improves all three metrics (\methodbase{}), and optimizing the allocation improves all three again (\method{}). Replacing the primitive state by a derived field is thus not enough; the gain comes from carrying both and from allocating the bits between them.

The gains depend on the budget (Table~\ref{tab:suite}b). At $R=0.25$, \method{} lowers $\nrmse$ by 15.5\% relative to \texttt{Primitive} and by 7.5\% relative to \texttt{ArchMulti}, and it raises $\horizon$ from 0.127 and 0.316 to 0.446. At $R=0.5$, its $\nrmse$ is 7.2\% below that of \texttt{ArchMulti} in aggregate (paired difference $-0.0047$ $[-0.0104,-0.0002]$); the gain comes from diffusion--reaction, INS, and compressible Navier--Stokes, while the other four families select the primitive encoding, and \method{} has the lower $\nrmse$ in only 36 of the 84 configurations at this budget, so the improvement there is an aggregate one. At $R=1$, the selected state is the primitive state, the gain from state design is zero, and the small lead of \texttt{ArchMulti} reflects its predictor. The rollout gain thus follows the input deficit of Section~\ref{sec:mechanism}: it is largest where the primitive state loses most at input and vanishes once the budget suffices for that state. Under a fixed budget, the fine-scale loss of the carried state therefore persists into the rollout, and \method{}, which reduces this loss at its source, outperforms the predictor-side and codec-side alternatives at the same budget in both fine-band error and detail-valid horizon.

\subsubsection{Generality, complementarity, and stability}\label{sec:additional}

\paragraph{No equation-level knowledge.} We replace the seven expert libraries by one shared library generated by a frozen rule from tensor type, dimension, and boundary type alone: each primitive scalar, four spectral powers of the boundary-adapted Laplacian, and four band projections, plus divergence and curl for vector fields and a DCT-II rotation across the components of multichannel states (120 channels, 108 of them derived; Appendix~\ref{app:libraries}). In the canonical regime with FNO, \method{} selects the same effective encodings as with the expert libraries and assigns zero bits to 100 of the 108 derived channels, so all results are unchanged. Averaged over the seven families, \method{} reaches $\nrmse$ 0.0511 and $\horizon$ 0.486, against 0.0602 and 0.162 for \texttt{Primitive}, 0.0576 and 0.256 for \bestsingle{}, and 0.0563 and 0.342 for \methodbase{} under the same library. The gains therefore do not depend on hand-picking a winning field per PDE.

\paragraph{Non-uniform grids.} On the CylinderFlow meshes of MeshGraphNets~\cite{pfaff2020meshgraphnets}, an input-stage mesh protocol keeps 25\% of the nodes (e.g., 1,888 to 472) at 16 bits per retained node, i.e., 944~B per state. \method{} selects $u$\,5+5, $\omega$\,6 and lowers the input $\finerel(0)$ from 0.992 (\texttt{Primitive}, shared decoder) to $0.649\pm0.015$ over five runs, so the input-stage gain does not require a uniform grid. We evaluate this case at the input stage only.

\paragraph{Complementarity.} Feeding the \method{} state to \texttt{ArchMulti} requires no change to either method. On INS, \texttt{DerivOpt+ArchMulti} improves on both single methods on each of the four backbones, with macro-average $\nrmse$ 0.0703 and $\horizon$ 0.522, so state design adds to predictor design rather than replacing it.

\paragraph{Stability and selected designs.} \method{} has the highest $\horizon$ in 130 of the 252 configurations (10 of them ties). It keeps the highest pooled $\horizon$ under stricter validity thresholds ($\qfine\in[1/1.15,1.15]$, $\eout<0.075$), looser ones ($\qfine\in[1/1.5,1.5]$, $\eout<0.15$), and a wider fine band $(2k_c/3,k_c)$ (Appendix~\ref{app:extra}), so its horizon gain does not hinge on these choices. Of the 63 selected designs, 30 carry derived fields: all 21 at $R=0.25$, 9 at $R=0.5$, and none at $R=1$; the tighter the budget, the more often derived fields are worth their bits. Appendix~\ref{app:extra} lists the canonical designs.

\section{Discussion}\label{sec:discussion}

We showed that under a fixed storage budget, a neural PDE surrogate first loses fine-scale information in its carried state, and that how much is lost depends on which fields are carried and how their bits are allocated. Choosing them with \method{} makes more of this information recoverable from the same bits: the fine-band error falls already at the input and stays lower in rollouts, which reach the longest detail-valid horizons and the lowest pooled full-field error across seven PDE families and four backbones; the gain adds to that of a multiscale architecture, and the primitive state is kept once the budget suffices for it. Because no predictor can reliably recover sample-specific detail that the carried state no longer contains, the representation supplied to a learned simulator is part of the learning problem, and carried-state design should be treated as a first-class design axis alongside architecture, loss, and rollout strategy rather than as a fixed backdrop to predictor design.

Notably, this axis gains weight as surrogates scale. A 500-step trajectory of velocity and pressure on a $1024^3$ grid occupies 7.81~TiB in FP32, and a single width-20 FNO feature map at that resolution needs 80~GiB, the entire memory of an 80~GB GPU. Coarsening by eight per axis brings the feature map to 160~MiB, and storing the coarsened state at 8 bits per component brings the trajectory to 3.91~GiB; these are the budget ratio and retained fraction of the canonical regime of Section~\ref{sec:empirical}, at which the primitive INS state had already lost its top retained band. At such scales, reducing the state is a precondition for training rather than an option, so the choice of carried fields decides how much fine-scale physics a surrogate can learn at all. Numerical methods have long chosen their prognostic variables for the structure they preserve; for learned simulators under a bit budget, \method{} makes the same choice by optimization, before any network is trained, for any predictor, and without knowledge of the governing equation.

The main limitation is that \method{} selects among candidate fields but does not construct them. Its improvement over the best primitive-only design requires the library to contain at least one field that preserves the target band better per bit than the primitive state; otherwise it returns that design, and calibration and selection add cost without gain. What matters is therefore the coverage of the library, whereas redundancy does not hurt the calibrated optimum (Proposition~\ref{prop:existence}) and only enlarges the search. A library generated without equation-level knowledge sufficed in our experiments (Section~\ref{sec:additional}), but this does not guarantee a useful candidate for every PDE; for less understood dynamics, candidates may have to come from operator libraries built or learned from data, such as PDE-FIND and PDE-Net~\cite{rudy2017pdefind,long2018pdenet}, and learning candidates jointly with the selection remains open.

The selected design also depends on calibrated transfer and residual statistics under a linear-Gaussian channel model: if the calibration data do not represent the deployment regime or the model is misspecified, the selected subset and allocation can change, and a lower calibrated risk need not give better rollouts. The size and spectral distribution of the input loss depend on the storage operator; our experiments cover one family of storage operators, basis truncation followed by uniform scalar quantization, and one learned codec, and other coarseners, quantizers, and codecs require recalibration and were not tested. The claims concern explicit-state autoregressive surrogates under tight budgets and do not imply that all fine-scale error of neural simulators originates in the carried state; outside this setting, model expressiveness can remain the dominant bottleneck. Carrying derived fields also adds per-step encoding and fusion work, the non-uniform-mesh case was evaluated at the input stage only, and 3D benchmarks remain future work.

\begin{ack}
This work is supported by the National Natural Science Foundation of China (Grant No. 62372409) and the Ministry of Science and Technology of the People's Republic of China (Grant No. 2023ZD0120704 of Project No. 2023ZD0120700). The authors declare no competing interests.
\end{ack}

\section*{References}

\begingroup
\small
\renewcommand{\section}[2]{}

\endgroup

\appendix
\newpage

\section*{Contents of Appendices}
\vspace{1em}
\noindent
\hyperref[app:theory]{A \quad Derivations \dotfill \pageref{app:theory}}\\[0.5em]
\hyperref[app:operator]{B \quad State Construction, Calibration, and Selection \dotfill \pageref{app:operator}}\\[0.5em]
\hyperref[app:family_inst]{C \quad Family-Specific Instantiations \dotfill \pageref{app:family_inst}}\\[0.5em]
\hyperref[app:protocol]{D \quad Experimental Protocol \dotfill \pageref{app:protocol}}\\[0.5em]
\hyperref[app:extra]{E \quad Additional Results \dotfill \pageref{app:extra}}\\[0.5em]
\hyperref[sec:llm_usage]{F \quad Use of Large Language Models \dotfill \pageref{sec:llm_usage}}

\newpage

\section{Derivations}\label{app:theory}

\subsection{Primitive-state distortion (Section~\ref{sec:primitive})}\label{app:theory_primitive}

We expand the steps leading to Eq.~\eqref{eq:sec2_master_lb}. Let $u_j$ be the $j$-th velocity component, $\Lambda_f$ the fine Fourier lattice, $\Lambda_c:=\Lambda_f\cap\mathcal B_{\mathrm{exp}}$ the retained lattice, and $P_{\mathrm{exp}}$ the spectral projector onto $\Lambda_c$. Primitive coarsening band-limits the field, stores coarse nodal values, and quantizes them componentwise: for $x_n^c:=\frac{2\pi n}{N_c}$ and $j=1,\ldots,d$,
\begin{equation}
 u_j^{\mathrm{lp}}=P_{\mathrm{exp}}u_j,
 \qquad
 u_j^c[n]=u_j^{\mathrm{lp}}(x_n^c),
 \qquad
 y_j[n]=Q_{b_j}(u_j^c[n]),
 \qquad
 \sum_{j=1}^d b_j=B.
\label{eq:appA_samples}
\end{equation}
Because $u_j^{\mathrm{lp}}$ is band-limited to $\Lambda_c$, coarse trigonometric interpolation reconstructs it exactly before quantization. With $\eta_j[n]:=y_j[n]-u_j^c[n]$ and $\eta_j(x)$ its coarse trigonometric interpolant, the decoded field is $\tilde u_j=u_j^{\mathrm{lp}}+\eta_j$, so
\begin{equation}
 \widehat{\tilde u_j}(k)-\hat u_j(k)=\hat\eta_j(k),
 \qquad k\in\Lambda_c,
\label{eq:appA_error_identity}
\end{equation}
because $P_{\mathrm{exp}}$ preserves $\hat u_j$ on $\Lambda_c$ and removes every other mode; in particular, no folded alias term enters $\mathcal B_{\mathrm{hf}}$.

\paragraph{Noise and signal energy in the top band.} A uniform $b$-bit quantizer on the clipped standardized range $[-a,a]$ has step $2a\,2^{-b}$ and, in the high-rate regime, per-sample error variance $c_q2^{-2b}\sigma_j^2$ with $c_q=\frac{a^2}{3}$, where $\sigma_j^2$ is the variance of the stored component. Under the high-rate closure the errors are zero-mean and uncorrelated with the signal, across components, and across sites. With the unitary coarse discrete Fourier transform, white noise of per-sample variance $v$ has expected energy $v$ on each of the $N_c^d$ coarse modes, so
\begin{equation}
 \sum_{k\in\mathcal B_{\mathrm{hf}}}\mathbb E|\hat\eta_j(k)|^2
 =
 |\mathcal B_{\mathrm{hf}}|\,c_q2^{-2b_j}\sigma_j^2
 =
 N_c^d\,\vartheta_{\mathrm{hf}}\,c_q2^{-2b_j}\sigma_j^2,
 \qquad
 \vartheta_{\mathrm{hf}}:=\frac{|\mathcal B_{\mathrm{hf}}|}{N_c^d}.
\label{eq:appA_noise_band}
\end{equation}
Counting lattice points by volume, $\vartheta_{\mathrm{hf}}\approx 1-\gamma$ in 1D, $\frac{\pi}{4}(1-\gamma^2)$ in 2D, and $\frac{\pi}{6}(1-\gamma^3)$ in 3D. For a zero-mean component, the same transform gives $\sum_{k\in\Lambda_c}\mathbb E|\hat u_j(k)|^2=N_c^d\sigma_j^2$. Let
\begin{equation}
 \rho_{\mathrm{hf}}(r,\gamma)
 :=
 \frac{\sum_{k\in\mathcal B_{\mathrm{hf}}}\mathbb E\|\hat u(k)\|_2^2}{\sum_{k\in\mathcal B_{\mathrm{exp}}}\mathbb E\|\hat u(k)\|_2^2}
 \approx
 \frac{\int_{\gamma k_c}^{k_c}E(\kappa)\,d\kappa}{\int_{0}^{k_c}E(\kappa)\,d\kappa},
 \qquad k_c=\frac{N_f}{2r},
\label{eq:appA_rhohf}
\end{equation}
be the fraction of retained-band energy in $\mathcal B_{\mathrm{hf}}$, where $E(\kappa)$ is the shell spectrum. For isotropic flow every component has the same band fraction and the same variance $\sigma^2$. Inserting Eqs.~\eqref{eq:appA_error_identity}--\eqref{eq:appA_rhohf} into Eq.~\eqref{eq:sec2_hf_metric} gives
\begin{equation}
 \mathbb E\big[\mathrm{hf}_{L_2}(0)^2\big]
 =
 \frac{\sum_{j}N_c^d\vartheta_{\mathrm{hf}}c_q2^{-2b_j}\sigma^2}{d\,\rho_{\mathrm{hf}}N_c^d\sigma^2}
 =
 \frac{\vartheta_{\mathrm{hf}}c_q}{d\,\rho_{\mathrm{hf}}(r,\gamma)}
 \sum_{j=1}^d 2^{-2b_j}.
\label{eq:appA_quant_before_jensen}
\end{equation}
Because $b\mapsto 2^{-2b}$ is convex and $\sum_jb_j=B$, Jensen's inequality yields $\frac1d\sum_{j}2^{-2b_j}\ge2^{-\frac{2B}{d}}$, with equality at the equal split $b_1=\cdots=b_d=\frac Bd$. Therefore
\begin{equation}
 \mathbb E\big[\mathrm{hf}_{L_2}(0)^2\big]
 \ge
 \frac{\vartheta_{\mathrm{hf}}\,a^2}{3\,\rho_{\mathrm{hf}}(r,\gamma)}2^{-\frac{2B}{d}},
\label{eq:appA_master}
\end{equation}
which is Eq.~\eqref{eq:sec2_master_lb}. Anti-aliasing removes every unresolved mode, but quantization still injects approximately white error over all coarse modes, while the top retained band holds only the fraction $\rho_{\mathrm{hf}}$ of the retained energy. Its squared relative error therefore exceeds that of the field as a whole, $\frac{a^2}{3}2^{-2B/d}$, by the factor $\vartheta_{\mathrm{hf}}/\rho_{\mathrm{hf}}$, which grows as the spectrum steepens; the budget $B$ sets the overall level. Figure~\ref{fig:sec2_phase} evaluates Eq.~\eqref{eq:appA_master} in the canonical INS setting of Section~\ref{sec:empirical}: $N_f=256$, the top band $(0.75\,k_c,k_c)$ of its metrics ($\gamma=0.75$), the clip $a=4$ of Appendix~\ref{app:operator}, and $b$ bits per velocity component, $B=2b$. The INS spectrum is not available in closed form, so we use $E(\kappa)=1/(1+(\kappa/4)^{\alpha})$ with $\alpha=9.13$, the value at which Eq.~\eqref{eq:appA_master}, with $\vartheta_{\mathrm{hf}}\approx\frac{\pi}{4}(1-\gamma^2)$ and $\rho_{\mathrm{hf}}$ from the shell integral, equals the measured direct-decode input error, 1.394, at $B=16$ and $r=8$ (Section~\ref{sec:empirical}). Figure~\ref{fig:sec3_field_distortion} uses the same spectrum at $r=8$.

\subsection{Field-dependent distortion (Section~\ref{sec:fielddist})}\label{app:theory_fielddist}

In 2D incompressible flow, each Fourier mode has a single divergence-free latent scalar $z(k)$, with $\hat u(k)=e_{\perp}(k)z(k)$ and $\hat \omega(k)=i|k|z(k)$. For any candidate channel $c$ we write
\begin{equation}
 \hat x_c(k)=M_c(k)z(k),
\label{eq:appA_channel_symbol}
\end{equation}
so that $M_u(k)=1$ after projection onto the divergence-free direction and $M_{\omega}(k)=i|k|$. Applying the operator of Section~\ref{sec:primitive} channel-wise gives
\begin{equation}
 \widehat{\tilde x_c}(k)
 =
 M_c(k)z(k)+\hat\eta_c(k),
 \qquad |k|<k_c,
\label{eq:appA_channel_alias}
\end{equation}
where $\hat\eta_c(k)$ is the quantization noise of channel $c$ on the retained band. Dividing by $M_c(k)$ yields $\hat z_c(k)-z(k)=\hat\eta_c(k)/M_c(k)$, and under the same closure, with $S_z(k):=\mathbb E|z(k)|^2$ and $\sigma_c^q(k;b_c):=\mathbb E|\hat\eta_c(k)|^2$,
\begin{equation}
 \mathcal D_c(k)
 :=
 \frac{\mathbb E|\hat z_c(k)-z(k)|^2}{S_z(k)}
 =
 \frac{\sigma_c^q(k;b_c)}{|M_c(k)|^2S_z(k)},
\label{eq:appA_field_master}
\end{equation}
which is Eq.~\eqref{eq:sec3_master_curve}; for velocity and vorticity it gives Eq.~\eqref{eq:sec3_uomega_curves}. The shell curves of Figure~\ref{fig:sec3_field_distortion} aggregate Eq.~\eqref{eq:appA_field_master} over the shells $\mathcal S_{\kappa}:=\{k:\kappa-\frac12<|k|\le\kappa+\frac12\}$:
\begin{equation}
 L2_c(\kappa)^2
 :=
 \frac{\sum_{k\in\mathcal S_{\kappa}}\mathbb E|\hat z_c(k)-z(k)|^2}
 {\sum_{k\in\mathcal S_{\kappa}}S_z(k)}
 =
 \frac{\sum_{k\in\mathcal S_{\kappa}}S_z(k)\mathcal D_c(k)}
 {\sum_{k\in\mathcal S_{\kappa}}S_z(k)}.
\label{eq:appA_shell_curve}
\end{equation}
If the shell-wise quantization spectra vary slowly, then $\mathcal D_{\omega}(k)/\mathcal D_u(k)\approx\sigma_{\omega}^q(\sigma_u^q)^{-1}|k|^{-2}$, so the vorticity curve falls relative to the velocity curve as $|k|$ approaches the cutoff, and a crossover appears near $|k|_{\times}\approx(\sigma_{\omega}^q/\sigma_u^q)^{1/2}$ whenever this variance ratio is approximately constant.

\subsection{Posterior risk and the Navier--Stokes specialization (Section~\ref{sec:derivopt})}\label{app:theory_derivopt}

For each selected channel $c\in S$, the decoded observation is modeled as $\hat y_c(k)=B_c(k)M_c(k)z(k)+n_c(k)$, where $z(k)\in\mathbb C^{d-1}$ is the divergence-free latent state and $n_c(k)$ has covariance $\Sigma_c(k;b_c)$. Stacking the selected channels gives
\begin{equation}
 \hat y_S(k)=A_S(k)z(k)+n_S(k),
 \qquad
 A_S(k):=\operatorname{rowstack}_{c\in S}[B_c(k)M_c(k)],
\label{eq:appA_stacked_model}
\end{equation}
with covariance $\Sigma_S(k;\mathbf b)$ for $n_S(k)$, and we use the prior $z(k)\sim\mathcal{CN}(0,S_z(k)I_{d-1})$. The system is linear and Gaussian, so the posterior is Gaussian with
\begin{equation}
\begin{aligned}
 \Sigma_{\mathrm{post},S}(k)
 &=
 \big(S_z(k)^{-1}I_{d-1}+A_S(k)^\ast\Sigma_S(k;\mathbf b)^{-1}A_S(k)\big)^{-1},\\
 \hat z_S(k)
 &=
 \Sigma_{\mathrm{post},S}(k)A_S(k)^\ast\Sigma_S(k;\mathbf b)^{-1}\hat y_S(k).
\end{aligned}
\label{eq:appA_post_cov}
\end{equation}
The posterior mean $\hat z_S$ is the shared decoder. For a target $\hat g(k)=L(k)z(k)$, the Bayes risk of the selected state is
\begin{equation}
 \mathcal J(S,\mathbf b)
 =
 \sum_{k\in\mathcal B_{\star}}w(k)
 \operatorname{tr}\big(L(k)\Sigma_{\mathrm{post},S}(k)L(k)^\ast\big),
\label{eq:appA_general_score}
\end{equation}
which is Eq.~\eqref{eq:sec4_posterior} on a Fourier basis. For incompressible Navier--Stokes the target is velocity. In 2D, $\hat u(k)=e_{\perp}(k)z(k)$ with $e_{\perp}^\ast e_{\perp}=1$; in 3D, $\hat u(k)=E(k)z(k)$ with $E^\ast E=I_2$. The velocity map is thus an isometry of the latent variable, and
\begin{equation}
 \mathcal J(S,\mathbf b)
 =
 \sum_{k\in\mathcal B_{\star}}w(k)\,\operatorname{tr}\Sigma_{\mathrm{post},S}(k).
\label{eq:appA_ns_reduction}
\end{equation}
For $\mathcal C=\{u,\omega\}$, $M_u=e_\perp$ and $M_\omega=i|k|$ in 2D, and $M_u=E$ and $M_\omega=i|k|G$ in 3D, where $G(k)$ is the orthonormal curl basis with $G^\ast G=I_2$. With isotropic residuals $\Sigma_u=\sigma_u^{\mathrm{eff}}I_d$ and $\Sigma_\omega=\sigma_\omega^{\mathrm{eff}}I_d$, the channel precisions add as multiples of $I_{d-1}$,
\begin{equation}
 A_S^\ast\Sigma_S^{-1}A_S
 =
 \Big(
 \mathbf 1_{\{u\in S\}}\frac{|B_u(k)|^2}{\sigma_u^{\mathrm{eff}}(k;b_u)}
 +
 \mathbf 1_{\{\omega\in S\}}\frac{|B_{\omega}(k)|^2|k|^2}{\sigma_{\omega}^{\mathrm{eff}}(k;b_{\omega})}
 \Big)I_{d-1},
\label{eq:appA_precision_add}
\end{equation}
and substituting into Eqs.~\eqref{eq:appA_post_cov} and~\eqref{eq:appA_ns_reduction} gives Eq.~\eqref{eq:sec4_ns_objective}.

\subsection{Proof of Proposition~\ref{prop:existence} and an extended library}\label{app:proof}

\paragraph{(i) Existence and exactness.} For $c\in S$, $b_c\in\{1,\ldots,\lfloor B/m_c\rfloor\}$. There are at most $2^{|\mathcal C|}$ subsets and finitely many allocations per subset, so $|\mathcal D_B|<\infty$, and $\mathcal D_B\neq\emptyset$ because a primitive channel $c$ with $m_c\le B$ is feasible at one bit per component. For every $(S,\mathbf b)\in\mathcal D_B$ and every $\ell$, the matrix $M_S(\ell):=S_z(\ell)^{-1}+A_S(\ell)^\ast\Sigma_S(\ell;\mathbf b)^{-1}A_S(\ell)$ is Hermitian with $M_S(\ell)\succeq S_z(\ell)^{-1}\succ0$, hence invertible, and $\operatorname{tr}(L_gM_S^{-1}L_g^\ast)\in[0,\infty)$. Thus $\mathcal J$ is a real-valued function on a finite nonempty set and attains its minimum. Exhaustive enumeration returns a minimizer, and so does any search that discards a design only when a valid lower bound on its risk exceeds the risk of a feasible incumbent (Appendix~\ref{app:selection}).

\paragraph{(ii) Primitive fallback.} Let $\mathcal D_B^{\mathcal P}:=\{(S,\mathbf b)\in\mathcal D_B:S\subseteq\mathcal P\}$. Since $\mathcal D_B^{\mathcal P}\subseteq\mathcal D_B$, $\min_{\mathcal D_B}\mathcal J\le\min_{\mathcal D_B^{\mathcal P}}\mathcal J$. Equality holds exactly when the minimum over $\mathcal D_B$ is attained in $\mathcal D_B^{\mathcal P}$, that is, when no design that carries a non-primitive channel has lower risk.

\paragraph{(iii) Library monotonicity.} The calibrated channel of a candidate does not depend on which other candidates are in the library, so $\mathcal D_B(\mathcal C)\subseteq\mathcal D_B(\mathcal C')$ with identical risks on the common designs, and $\min_{\mathcal D_B(\mathcal C')}\mathcal J\le\min_{\mathcal D_B(\mathcal C)}\mathcal J$.

\paragraph{Extended library.} Any finite set of linear candidates fits the same template. In periodic 2D incompressible flow, consider
\begin{equation}
 \mathcal C_{\mathrm{ext}}=\{u,\ \omega,\ \psi,\ \nabla u,\ \varepsilon(u),\ \nabla\cdot u,\ \Delta u,\ \Lambda u\},
 \qquad
 \varepsilon(u)=\tfrac12\big(\nabla u+(\nabla u)^\top\big),
 \qquad
 \Lambda=(-\Delta)^{1/2},
\label{eq:appA_ext_library}
\end{equation}
where $\psi$ is the streamfunction. Each candidate has a Fourier symbol $M_c(k)$ acting on the scalar latent $z(k)$, with $\mu_c(k):=M_c(k)^\ast M_c(k)$ equal to $1$, $|k|^2$, $|k|^{-2}$, $|k|^2$, $|k|^2/2$, $0$, $|k|^4$, and $|k|^2$, respectively. With isotropic residuals, the precisions add and
\begin{equation}
 \mathcal J_{\mathrm{ext}}(S,\mathbf b)
 =
 \sum_{k\in\mathcal B_\star}w(k)\Big(S_z(k)^{-1}+\sum_{c\in S}\frac{|B_c(k)|^2\mu_c(k)}{\sigma_c^{\mathrm{eff}}(k;b_c)}\Big)^{-1}.
\label{eq:appA_ext_score}
\end{equation}
The divergence of a divergence-free field has $\mu=0$: it adds cost but no information, so a minimizer with ties broken toward lower cost never carries it. The extended problem has the same structure and the same exact solution procedure as the two-field case; only $\mathcal D_B$ is larger.

\section{State Construction, Calibration, and Selection}\label{app:operator}

\subsection{Coarsen--quantize--decode operator}\label{app:coarsen_protocol}

Every explicit-state method uses the same construction:
\begin{enumerate}
    \item compute each carried field from the primitive state;
    \item project it onto the retained band of the family's basis (strict truncation, $|k|<k_c$ in the periodic case) and sample it on the $n_r$ retained sites;
    \item standardize each stored component with a mean and scale frozen from training data, clip it to $\pm4$ standard deviations, and quantize it uniformly with $b_c$ bits;
    \item serialize the codes; this payload is the external state passed between autoregressive steps;
    \item decode by dequantizing and fusing all stored channels with the posterior mean of Eq.~\eqref{eq:appA_post_cov}, which returns the coarse primitive field given to the predictor.
\end{enumerate}
Direct decoding, which dequantizes and interpolates the primitive channels without fusion, is used only as a diagnostic (Section~\ref{sec:results}).

\subsection{Calibration}\label{app:calibration}

Calibration uses training trajectories only. Passing training states through the operator, it estimates the prior $S_z(\ell)$ and, for every candidate and admissible width, the in-band transfer $B_c(\ell)$ and residual covariance $\Sigma_c(\ell;b_c)$ of Eq.~\eqref{eq:sec4_channel}. Statistics are pooled over shells of similar frequency and keep the full covariance across stored components. Candidates evaluated by physical operators rather than spectral symbols (finite-difference gradients on non-periodic grids and the Burgers flux) enter through training-fitted effective responses. Widths are capped at 26 bits per component, an empirical precision plateau of our float32 pipeline, and the 4-standard-deviation clip is likewise an empirical choice; at $R=1$ the selected designs use 26 bits per primitive component and pad the remaining bits.

\subsection{Exact selection}\label{app:selection}

For each candidate the admissible options are ``absent'' and every calibrated width that fits the budget; an option whose information matrix equals that of a cheaper option is discarded. The search seeds incumbents with all single-candidate designs and with greedy designs, orders candidates by their best single-candidate risk, and visits the tree of options depth first. A node is pruned only if one of three valid lower bounds on $\mathcal J$ over all of its completions exceeds the incumbent risk: a positive-semidefinite envelope of the remaining options, a multiple-choice knapsack bound on the diagonal of the information matrix, and a supporting-plane bound from the convexity of $\mathcal J$ in the information matrix. The returned design is therefore a global minimizer of Eq.~\eqref{eq:sec4_problem}. Ties are broken by lower cost and then by the lexicographically smaller allocation; no validation or test metric is consulted. One selection per family and regime (63 runs) serves all four backbones. \bestsingle{} is the single derived channel with the lowest calibrated risk at the full budget, and \methodbase{} keeps the subset selected by \method{} with an equal split.

\subsection{Candidate libraries}\label{app:libraries}

Table~\ref{tab:app_libraries} lists the expert libraries. $\Lambda=(-L)^{1/2}$ for the family's boundary-adapted Laplacian $L$ (periodic, Robin/Cauchy-adapted $L_R$, or Neumann-adapted $\Delta_N$). The shared library of Section~\ref{sec:additional} is generated by one frozen rule that uses only the role of each primitive component (scalar or vector), the spatial dimension, and the boundary type. Each primitive scalar $x$ contributes nine channels: $x$; $(-L)^sx$ for $s\in\{\frac12,1,\frac32,2\}$; and the projections of $x$ onto the radial bands $[0,0.25)$, $[0.25,0.5)$, $[0.5,0.75)$, and $[0.75,1]$ of the retained cutoff. Each vector field also contributes its divergence and curl, and each multichannel state a DCT-II rotation across its physical components. Every generated channel is RMS-normalized on the training split. The rule never uses the identity, equation, or coefficients of the PDE or any validation or test result, and no expert field is inserted by name.

\begin{table}[h]
    \centering
    \small
    \setlength{\tabcolsep}{4pt}
    \caption{Expert candidate libraries and library sizes in scalar channels (expert / shared).}
    \label{tab:app_libraries}
    \begin{tabular}{llll}
    \toprule
    \textbf{Family} & \textbf{Primitive} & \textbf{Expert library} & \textbf{Channels} \\
    \midrule
    Advection & $u$ & $u,\ \Lambda u,\ -\Delta u,\ u_x,\ u_{xx}$ & 5 / 9 \\
    Burgers & $u$ & $u,\ \Lambda u,\ -\Delta u,\ \frac{u^2}{2},\ \partial_x\frac{u^2}{2}$ & 5 / 9 \\
    Diffusion--sorption & $u$ & $u,\ \Lambda_R u,\ -L_R u,\ \partial_x u$ & 4 / 9 \\
    Diffusion--reaction & $q=(u,v)$ & $q,\ \Lambda_N q,\ -\Delta_N q,\ d=\frac{u-v}{\sqrt2},\ s=\frac{u+v}{\sqrt2}$ & 8 / 20 \\
    Radial dam break & $h$ & $h,\ \Lambda_N h,\ -\Delta_N h,\ \partial_r h$ & 4 / 9 \\
    Incompressible NS & $u$ & $u,\ \omega,\ \psi,\ \Lambda u$ & 6 / 22 \\
    Compressible NS & $q=(\rho,v,p)$ & $q,\ \chi=\nabla\cdot v,\ \omega,\ \Lambda p,\ \Lambda\rho,\ -\Delta p$ & 9 / 42 \\
    \midrule
    Total & & & 41 / 120 \\
    \bottomrule
    \end{tabular}
\end{table}

\section{Family-Specific Instantiations}\label{app:family_inst}

Each family instantiates Eqs.~\eqref{eq:sec4_channel}--\eqref{eq:sec4_posterior} with a basis $\{\phi_\ell\}$ that satisfies its boundary conditions, the eigenvalues $\lambda_\ell\ge0$ of the corresponding negative Laplacian, $-L\phi_\ell=\lambda_\ell\phi_\ell$, and the primitive field as target.

\paragraph{Scalar families.} Advection and Burgers use the periodic Fourier basis, diffusion--sorption a Robin/Cauchy-adapted basis, and radial dam break a Neumann-adapted basis; the latent coefficient is $z(\ell)=\langle x,\phi_\ell\rangle$ for the primitive scalar $x\in\{u,h\}$. A spectral candidate $x_c=(-L)^{t_c}x$ has symbol $M_c(\ell)=\lambda_\ell^{t_c}$, with $t_c=0,\frac12,1$ for $x$, $\Lambda x$, and $-Lx$. With scalar residual variances the score is
\begin{equation}
 \mathcal J_{\mathrm{scalar}}(S,\mathbf b)
 =
 \sum_{\ell\in\mathcal I_\star}w_\ell\Big(S_x(\ell)^{-1}+\sum_{c\in S}\frac{|B_c(\ell)|^2\lambda_\ell^{2t_c}}{\sigma_c^2(\ell;b_c)}\Big)^{-1}.
\label{eq:appC_scalar}
\end{equation}
Physical candidates (finite-difference gradients and the Burgers flux) enter through their fitted responses $B_c(\ell)M_c(\ell)$.

\paragraph{Diffusion--reaction.} The primitive state is $q=(u,v)^\top$. The common and imbalance modes $s=\frac{u+v}{\sqrt2}$ and $d=\frac{u-v}{\sqrt2}$ are expanded in a Neumann-adapted basis, $z(\ell)=[\hat s(\ell),\hat d(\ell)]^\top$, and $R$ denotes the fixed rotation between $(u,v)$ and $(s,d)$. The posterior precision is
\begin{equation}
\resizebox{0.9\linewidth}{!}{$\displaystyle
\Sigma_{\mathrm{post}}^{-1}(\ell)=S_z(\ell)^{-1}
+\mathbf 1_{\{q\in S\}}R^*\frac{|B_q(\ell)|^2}{\sigma_q^2(\ell;b_q)}R
+\mathbf 1_{\{\Lambda_N q\in S\}}R^*\frac{|B_{\Lambda}(\ell)|^2\lambda_{\ell}}{\sigma_{\Lambda}^2(\ell;b_{\Lambda})}R
+\mathbf 1_{\{\Delta_N q\in S\}}R^*\frac{|B_{\Delta}(\ell)|^2\lambda_{\ell}^2}{\sigma_{\Delta}^2(\ell;b_{\Delta})}R
+\mathbf 1_{\{d\in S\}}e_2\frac{|B_d(\ell)|^2}{\sigma_d^2(\ell;b_d)}e_2^\top,
$}
\label{eq:appC_dr}
\end{equation}
with the analogous term for $s$, and the risk is $\sum_\ell w_\ell\operatorname{tr}(R\,\Sigma_{\mathrm{post}}(\ell)R^\top)$.

\paragraph{Incompressible Navier--Stokes.} The periodic specialization is Eq.~\eqref{eq:sec4_ns_objective}. \textsc{PDEBench} INS is wall-bounded, so its basis is boundary-adapted and the curl and streamfunction candidates are evaluated by physical operators with fitted responses; the score is the general form Eq.~\eqref{eq:sec4_posterior}.

\paragraph{Compressible Navier--Stokes.} The primitive state is $q=(\rho,v,p)$, with a Helmholtz-adapted latent decomposition $z=(\rho,\phi,\psi,p)$, $v=\nabla\phi+\nabla^{\perp}\psi$. With $L_q(\ell)$ mapping the latent variables to the primitive target and $A_S(\ell)$ stacking the selected calibrated channels,
\begin{equation}
\begin{aligned}
\Sigma_{\mathrm{post}}(\ell)&=\Big(S_z(\ell)^{-1}+A_S(\ell)^*\Sigma_S(\ell;\mathbf b)^{-1}A_S(\ell)\Big)^{-1},\\
\mathcal J_{\mathrm{cNS}}(S,\mathbf b)&=\sum_{\ell\in\mathcal I_{\star}}w_{\ell}\,\mathrm{tr}\big(L_q(\ell)\Sigma_{\mathrm{post}}(\ell)L_q(\ell)^*\big).
\end{aligned}
\label{eq:appC_cns}
\end{equation}

\section{Experimental Protocol}\label{app:protocol}

\paragraph{Data.} Reference grids have 1,024 points for the three 1D families, $128^2$ for diffusion--reaction and radial dam break, and $256^2$ for INS; compressible Navier--Stokes is projected from $512^2$ to $128^2$ by a periodic Galerkin projection. INS rollouts start at frame 40 ($t=0.20$) and all other families at frame~0; every rollout then predicts all remaining stored frames, $h=1,\ldots,T$. Each family uses 1,000 trajectories split 640/104/256, and the test cohort is shared by all methods, budgets, and seeds. The \textsc{PDEBench} data are released under CC BY 4.0~\cite{takamoto2022pdebench}. CylinderFlow is obtained with the download script of the MeshGraphNets repository (Apache License 2.0)~\cite{pfaff2020meshgraphnets} and is evaluated at the input stage only, keeping 25\% of the mesh nodes at 16 bits per node (944~B per state).

\paragraph{Budget ledger.} With $n_r=(fN)^d$ retained sites and $m_p$ primitive scalar components, the cap is $\lfloor 32\,m_pRn_r/8\rfloor$ bytes. Every serialized component counts against it: field codes $n_r\sum_{c\in S}m_cb_c$ bits, latent primary and hyperprior codes, headers, and zero padding up to the cap. Field identities, widths, scales, and normalization statistics are fixed in the checkpoint and transmit no bits per state. In the INS canonical regime, $n_r=32^2$, $B=16$, and every explicit-state record is exactly 2,048~B: \texttt{Primitive}, \texttt{ArchMulti}, and \texttt{RolloutMulti} use $u$\,8+8, \bestsingle{} $\omega$\,16, \methodbase{} $u$\,5+5 and $\omega$\,5 plus one padding bit per site, and \method{} $u$\,4+4 and $\omega$\,8. \texttt{LatentAE-Hyper} is accounted by its realized bitstream with per-state rate control; on INS its unpadded payload averages 1,747.9~B (maximum 2,046~B) and is zero-padded to exactly 2,048~B. Architecture-internal tensors, such as the ConvLSTM hidden and cell states, are identical across methods and are not part of the payload; the cap equalizes external-state bits, not model size or computation.

\paragraph{Training.} Each method and configuration is tuned over learning rate $\{3\times10^{-4},10^{-3},3\times10^{-3}\}$, weight decay $\{0,10^{-4}\}$, and gradient clipping $\{0.5,1\}$ (12 trials of 10,000 updates) on the validation split with seed 11. The selected trial is reused as the seed-11 run, and four further seeds (29, 47, 71, 103) are trained with the same setting. All experiments ran on 8 NVIDIA A100 GPUs.

\paragraph{Metrics.}\label{app:metrics} For the primitive target $x_h$ at step $h$ and its prediction $\hat x_h$, with the basis-adapted projectors $P_{\exp}$ and $P_{\mathrm{fine}}$ of Section~\ref{sec:exp_setup},
\begin{equation}
\begin{aligned}
\exprrel(h)&=\frac{\|P_{\exp}(\hat x_h-x_h)\|_2}{\|P_{\exp}x_h\|_2},
&\qquad
\finerel(h)&=\frac{\|P_{\mathrm{fine}}(\hat x_h-x_h)\|_2}{\|P_{\mathrm{fine}}x_h\|_2},\\[2pt]
\qfine(h)&=\frac{\|P_{\mathrm{fine}}\hat x_h\|_2^2/\|P_{\exp}\hat x_h\|_2^2}{\|P_{\mathrm{fine}}x_h\|_2^2/\|P_{\exp}x_h\|_2^2},
&\qquad
\eout(h)&=\frac{\|(I-P_{\exp})\hat x_h\|_2^2}{\|P_{\exp}x_h\|_2^2}.
\end{aligned}
\label{eq:appD_metrics}
\end{equation}
A step is valid when $\exprrel(h)<1$, $\finerel(h)<1$, $\qfine(h)\in[\tau_Q^{-1},\tau_Q]$, and $\eout(h)<\tau_{\mathrm{out}}$, with $(\tau_Q,\tau_{\mathrm{out}})=(1.25,0.10)$. For each trajectory, $\horizon=\frac1T\max\{\tau\le T:\text{steps }1,\ldots,\tau\text{ are valid}\}$, with $\horizon=0$ if step 1 is invalid; the input $h=0$ is not part of the horizon, and trajectories are not filtered by their input. $\finerel(0)$, $\nrmse$, and mean $\finerel$ pool squared norms over the test trajectories (and steps) before taking ratios. Under strict truncation $\eout(0)$ is close to zero, and we use it only as a protocol check.

\paragraph{Canonical regime.}\label{app:canonical} The canonical regime $(R,f)=(0.25,1/8)$ is shared by all families and was selected once, before any test evaluation, on a pilot of 100 INS training trajectories rolled out with all four backbones, i.e., 400 equally weighted rollouts per regime. A non-finite rollout counts as final-step $\nrmse$ 1 and $\horizon=0$. A regime is eligible if at least 95\% of the \texttt{Primitive} rollouts are finite, the mean final-step $\nrmse$ of \texttt{Primitive} is at most 0.15 while its mean $\horizon$ is at most 0.10, and \methodbase{} or \texttt{ArchMulti} reaches a mean final-step $\nrmse$ of at most 0.15. Among the eligible regimes, we take the one with the smallest payload ratio $Rf^2$ relative to the full-resolution FP32 state, breaking ties by smaller $R$ and then smaller $f$. The rule thus picks the smallest payload at which the primitive state already has a short detail-valid horizon while full-field errors remain moderate. No metric of \method{}, no horizon of the controls, and no test result enters the selection.

\paragraph{Statistics.} For each seed, a pooled metric averages the configurations it covers; tables report the mean and sample standard deviation over the five seeds. Paired differences ($\method{}$ minus comparator) are resampled 12,000 times hierarchically over families, configurations within families, and paired seeds to obtain 95\% intervals.

\section{Additional Results}\label{app:extra}

\begin{table}[!ht]
    \centering
    \small
    \setlength{\tabcolsep}{5pt}
    \caption{Designs selected by \method{} for the seven \textsc{PDEBench} families in the canonical regime $(R,f)=(0.25,1/8)$, in bits per scalar component; none uses padding. CylinderFlow does not use the grid retained fraction $f$; it follows a separate input-stage mesh protocol that keeps 25\% of the nodes (e.g., 1,888 to 472) at 16 bits per node, i.e., 944~B per state.}
    \label{tab:app_designs}
    \begin{tabular}{lcl}
    \toprule
    \textbf{Family} & \textbf{Bits/site} & \textbf{Selected design} \\
    \midrule
    Advection & 8 & $u$\,3,\ $\Lambda u$\,5 \\
    Burgers & 8 & $u$\,3,\ $\Lambda u$\,5 \\
    Diffusion--sorption & 8 & $u$\,3,\ $\Lambda_R u$\,5 \\
    Diffusion--reaction & 16 & $u$\,4,\ $v$\,4,\ $d$\,8 \\
    Radial dam break & 8 & $h$\,3,\ $\Lambda_N h$\,5 \\
    Incompressible NS & 16 & $u$\,4+4,\ $\omega$\,8 \\
    Compressible NS & 32 & $\rho$\,4,\ $v$\,4+4,\ $p$\,4,\ $\chi$\,8,\ $\omega$\,8 \\
    \midrule
    CylinderFlow (input only) & 16 per node & $u$\,5+5,\ $\omega$\,6 \\
    \bottomrule
    \end{tabular}
\end{table}

\paragraph{Selected designs.} Table~\ref{tab:app_designs} lists the canonical designs. Over all 63 family--regime cells, 30 designs carry derived fields: all 21 at $R=0.25$, the 9 cells of diffusion--reaction, INS, and compressible Navier--Stokes at $R=0.5$, and none at $R=1$. The other four families at $R=0.5$ select the 16-bit primitive encoding, and all families at $R=1$ select 26 bits per primitive component with padding, so \texttt{Primitive}, \methodbase{}, and \method{} coincide in those 33 cells. Bit widths do not simply scale with the budget. The designs of all 63 cells are listed in the code repository.

\begin{table}[!ht]
    \centering
    \small
    \setlength{\tabcolsep}{5pt}
    \caption{INS input stage, fine-band error by region (canonical regime, same test inputs). The near-wall region comprises the outermost 13 grid layers along all four walls on the $256\times256$ reference grid (zero-based indices with $\min(i,j,255-i,255-j)<13$), leaving a $230\times230$ interior; it contains 22.3\% of the true fine-band energy.}
    \label{tab:app_wall}
    \begin{tabular}{lcccc}
    \toprule
    \textbf{State} & \textbf{Global} & \textbf{Near-wall} & \textbf{Interior} & \textbf{Near-wall share of error energy} \\
    \midrule
    \texttt{Primitive} ($u$\,8+8) & 0.999 & 0.777 & 1.054 & 13.5\% \\
    \bestsingle{} ($\omega$\,16) & 0.863 & 1.300 & \textbf{0.689} & 50.5\% \\
    \methodbase{} ($u$\,5+5, $\omega$\,5) & 0.830 & 0.870 & 0.819 & 24.4\% \\
    \method{} ($u$\,4+4, $\omega$\,8) & \textbf{0.700} & \textbf{0.605} & 0.725 & 16.6\% \\
    \bottomrule
    \end{tabular}
\end{table}

\begin{table}[!ht]
    \centering
    \small
    \setlength{\tabcolsep}{5pt}
    \caption{Pooled horizon $\horizon$ over the 252 configurations under four evaluation protocols. The first three rows rescore the same predictions with different thresholds; the last rescores them with a wider fine band.}
    \label{tab:app_protocols}
    \begin{tabular}{lcccccc}
    \toprule
    \textbf{Protocol} & $(\tau_Q,\tau_{\mathrm{out}})$ & \textbf{Fine band} & \method{} & \texttt{ArchMulti} & \texttt{LatentAE-Hyper} \\
    \midrule
    Baseline & (1.25, 0.10) & $(0.75,1]$ & \textbf{0.524} & 0.458 & 0.463 \\
    Stricter & (1.15, 0.075) & $(0.75,1]$ & \textbf{0.329} & 0.287 & 0.290 \\
    Looser & (1.50, 0.15) & $(0.75,1]$ & \textbf{0.844} & 0.742 & 0.750 \\
    Wider band & (1.25, 0.10) & $(2/3,1]$ & \textbf{0.832} & 0.730 & 0.741 \\
    \bottomrule
    \end{tabular}
\end{table}

\paragraph{Wall partition and evaluation protocols.} Table~\ref{tab:app_wall} splits the INS input error of Section~\ref{sec:results} by region, and Table~\ref{tab:app_protocols} rescores the pooled horizon under stricter, looser, and wider-band protocols; \method{} has the highest $\horizon$ in every row.

\section{Use of Large Language Models}\label{sec:llm_usage}

LLMs were used only to assist with writing and presentation: checking technical terminology, correcting grammar, translating from the authors' native language, and improving sentence structure while preserving the scientific content. They were not used for research ideation, experimental design, data analysis, or any content that constitutes an intellectual contribution. All scientific insights, methodological decisions, and conclusions are the authors' own.

\end{document}